\documentclass{ecai}  

\usepackage{graphicx}
\usepackage{amssymb}
\usepackage{latexsym}

\ecaisubmission      

\paperid{3524}        
\usepackage{gensymb}
\usepackage{caption}
\usepackage{subcaption}
\begin{document}

\begin{frontmatter}

\title{Autonomous Navigation of Tractor-Trailer Vehicles through Roundabout Intersections}

\author[A]{\fnms{Daniel}~\snm{Attard}\thanks{Corresponding Author. Email: daniel.attard.18@um.edu.mt}}

\author[A]{\fnms{Josef}~\snm{Bajada}\orcid{0000-0002-8274-6177}}
\address[A]{Department of Artificial Intelligence, Faculty of ICT, University of Malta}

\begin{abstract}
In recent years, significant advancements have been made in the field of autonomous driving with the aim of increasing safety and efficiency. However, research that focuses on tractor-trailer vehicles is relatively sparse. Due to the physical characteristics and articulated joints, such vehicles require tailored models. While turning, the back wheels of the trailer turn at a tighter radius and the truck often has to deviate from the centre of the lane to accommodate this. Due to the lack of publicly available models, this work develops truck and trailer models using the high-fidelity simulation software Carla, together with several roundabout scenarios, to establish a baseline dataset for benchmarks. Using a twin-q soft actor-critic algorithm, we train a quasi-end-to-end autonomous driving model which is able to achieve a 73\% success rate on different roundabouts.
\end{abstract}

\end{frontmatter}

\section{Introduction}

Every year, approximately 1.3 million people die due to road traffic accidents with an additional 20 to 50 million people suffering non-fatal injuries \cite{world_health_organisation_road_2022}. With human error being the leading cause of road accidents, fully autonomous vehicles can help to reduce such fatalities significantly, by taking the human driver out of the equation. Furthermore, such vehicles can help with lowering emissions, increasing mobility, and reducing congestion.

Autonomous vehicle navigation is composed of four basic functions: localisation, perception, planning, and control \cite{Jo2014}. Despite some research focusing only on one function of autonomous vehicle navigation \cite{Capasso2020,Konstantinidis2021}, end-to-end systems \cite{chen_interpretable_2019,liu_efficient_2021} that combine all functions produce more generalised models that can handle more edge cases and unusual scenarios.

Despite significant research in this field, there is limited work publicly available that focuses on heavy goods vehicles. Apart from the aforementioned advantages, the application of autonomous navigation to heavy goods vehicles, specifically tractor-trailer vehicles, can lead to substantial economic growth by increasing the efficiency of the logistics sector, where human drivers are restricted by driving duration constraints and sleeping schedules. 


A tractor-trailer vehicle is defined as a truck with one trailer attached. The articulation point in such vehicles allows manoeuvring of tighter turns when compared to a single rigid vehicle of the same length, yet this introduces some undesirable behaviour. Firstly, when turning at lower speeds, the rear trailer wheels follow a tighter radius path than the front wheels of the truck \cite{kati_sadeghi_definitions_2013}. Secondly, due to their high centre of gravity, tractor-trailer vehicles are prone to rollover when turning at elevated speeds and tight turns. These physical properties need to be taken into consideration when designing autonomous driving models for such vehicles. In roundabout intersections, especially ones with a small radius, the truck has to deviate from the centre of the lane to have enough space for the trailer to turn safely without colliding with the kerb or rolling over. 

Several approaches have been proposed to achieve vehicle autonomy for small passenger vehicles. Traditional rule-based methods \cite{rastelli_fuzzy_2015} suffer from accuracy and applicability to real-life scenarios due to their hand-crafted nature. It is difficult to cover all possible scenarios, which leads to sub-optimal or erratic behaviour. Imitation Learning (IL) and Reinforcement Learning (RL) have also been explored as learning-based approaches. Given accurate training data, IL approaches \cite{liu_improved_2022} have shown to   achieve suitable performance. Despite the large amount of data required, they can only reach human level performance at best, assuming that the training data covers all the possible scenarios. On the other hand, RL algorithms are capable of surpassing such performance, converging towards an optimal policy  \cite{Capasso2020, chen_interpretable_2019,liu_improved_2022}. Online RL algorithms can learn from continually simulated data where different scenarios can be created, producing a more generic model. RL for autonomous driving can use various a combination of sensory inputs, which are not necessarily used by humans, such as images from different camera angles \cite{ Capasso2020,chen_interpretable_2019}, Light Detection and Ranging (LIDAR) \cite{cai_carl-lead_2021,liu_efficient_2021}, and vehicle information \cite{Konstantinidis2021}.

The research performed on heavy goods vehicles so far aimed to develop lane-changing models \cite{hoel_automated_2018, wang_intelligent_2022}, or used a low fidelity simulation with a very limited high-level action space that navigates through a roundabout and performs a right-hand turn \cite{Serin2020}. At the time of writing, such work focusing on a tractor-trailer vehicle navigating through a roundabout intersection with low-level longitudinal control is the first-of-its-kind.

One cause for the lack of research related to tractor-trailer vehicles is that very few high-fidelity models of such vehicles are publicly available. In light of this, the main contributions of this work are:
\begin{enumerate}
    \item Truck and trailer visual and physical models compatible with the high-fidelity, autonomous vehicle simulation software, Carla.
    \item A dataset of several roundabout scenarios, which can be used as a baseline for benchmarks.
    \item An RL environment that makes use of vehicle and route information, together with steering control actions.
    \item Preliminary results using a Twin-Q Soft Actor-critic (SAC) algorithm for a quasi-end-to-end autonomous driving model achieving a 73\% success rate.
\end{enumerate}




\begin{figure*}[htbp]
    \centering
\includegraphics[width=1\textwidth,keepaspectratio]{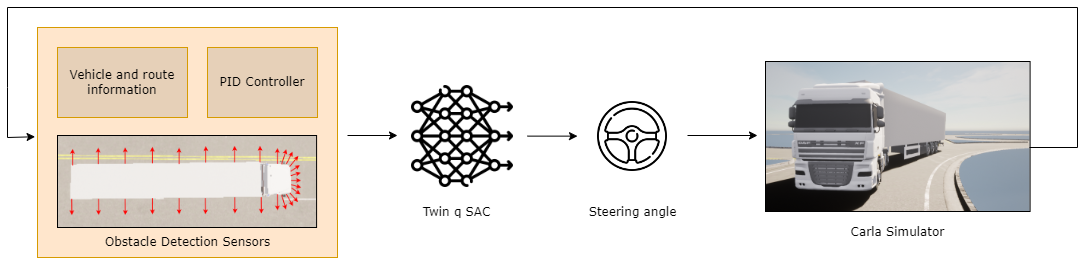}
    \caption[System Architecture]{An 
    overview of the system architecture
   }
   \label{fig:overview}
\end{figure*}

\section{Related Work}
\label{chp:related_work}
In recent years, there has been a significant amount of research on autonomous vehicles due to their rise in popularity. Furthermore, notable progress was made in deep reinforcement learning (DRL), which lead to its application to several fields, including autonomous vehicles. Some works propose a combination of autonomous navigation functions, developing end-to-end systems, while others solely focus on one specific function.

Q-learning, deep Q-Learning, and their variants have been widely used. Q-learning was utilised by Garcia Guenca et al. \cite{Cuenca2019b} to navigate a passenger vehicle through a multi-lane roundabout, controlling both steering and throttle values. Testing with and without traffic, their algorithm produced improved directionality and realistic average speed when compared to machine learning algorithms. Wolf et al. \cite{wolf_learning_2017} develop a lane-keeping system using a discrete action space and a deep Q-learning algorithm, achieving human-like behaviour. Converting Lidar data into an occupancy grid map, at 3 different intervals, Cai et al. \cite{cai_carl-lead_2021} develop an end-to-end autonomous driving system which controls the vehicle’s speed through an intersection. The dueling double deep q-learning algorithm achieved superior results when compared with baseline models, handling occlusion such as smaller vehicles hidden behind larger vehicles. The authors also noted that rule-based approaches tended to generate over-conservative models with a high success rate at the cost of completion time, while their RL algorithm achieved comparable results with significantly less completion time. Similarly, Chen et al. \cite{chen_interpretable_2019} developed their end-to-end system using bird-eye view images as input. The authors note that SAC performed superiorly when compared to double deep q-learning, TD3, and Deep Deterministic Policy Gradient (DDPG).

In order to reduce the problem's complexity and guarantee accurate steering, some research uses a pure-pursuit, path tracking, algorithm to handle steering control. Liu et al. \cite{liu_improved_2022} integrate the SAC and an imitation learning algorithm along with a bird’s-eye view image while controlling throttle and braking actions. Such integration combined with an improved prioritised replay buffer resulted in faster convergence rates while still producing an efficient model to manoeuvre a roundabout intersection. Similarly,  Zhang et al. \cite{Zhang2021} modify an actor-critic based RL algorithm to model the same task, obtaining excellent results.

Research in this field is mostly focused on smaller passenger vehicles. Very limited research was performed to develop a model to autonomously navigate heavy goods vehicles, which come with their own challenges, especially articulated vehicles. Using a rule-based algorithm called Lyapunov method, Widyotriatmo, Siregar and Nazaruddin \cite{widyotriatmo_line_2017} develop a line-following control system for a tractor-trailer vehicle using a predetermined path specifically designed for such articulated vehicles. Wang et al. \cite{wang_intelligent_2022} developed a high-fidelity truck and trailer model with realistic lateral and longitudinal kinematics, but at the time of writing, such model is not publicly available. Using vehicle information combined with a double deep q-network (DQN) algorithm, the authors produced a lane-changing model for a tractor-trailer vehicle using high-level actions. Serin \cite{Serin2020} compared several DQN based algorithms for manoeuvring a tractor-trailer vehicle through a roundabout intersection with oncoming traffic. Using SUMO as the simulation environment, only throttle actions were considered, since the vehicle followed a pre-planned route. 

As can be seen, very limited research deals with the autonomous navigation of articulated heavy goods vehicles. Furthermore, current research focuses on specific functions with high-level actions, sometimes lacking high fidelity simulations, which are required to correctly simulate the physical characteristics of these vehicles. In this work, we propose a high fidelity simulation environment, combined with a SAC RL algorithm to develop a quasi-end-to-end model to navigate a tractor-trailer vehicle around a roundabout intersection. Such intersections may have a tight radius forcing the vehicle to move away from the normal route.

\section{Methodology}
\label{chp:methodology}
This section will first give an overview of the whole system. Secondly, the development of the tractor-trailer model and roundabout scenarios will be described. Thirdly, the design of each RL component will be described in detail, justifying any decisions taken.

\subsection{System overview}

Figure \ref{fig:overview} illustrates an overview of the system. The Carla simulator \cite{dosovitskiy_carla_2017} is used to obtain observations of the current state of the vehicle through the provided Python API. These, along with the reward from the previous state, are fed through the RLlib \cite{liang_rllib_2018} framework to train the twin-q SAC algorithm. The SAC algorithm outputs the desired actions and such actions are performed in the simulation environment, progressing to the next set of observations.

\begin{figure*}[htbp]
     \centering
     \begin{subfigure}[b]{0.18\textwidth}
         \centering
         \includegraphics[width=\textwidth]{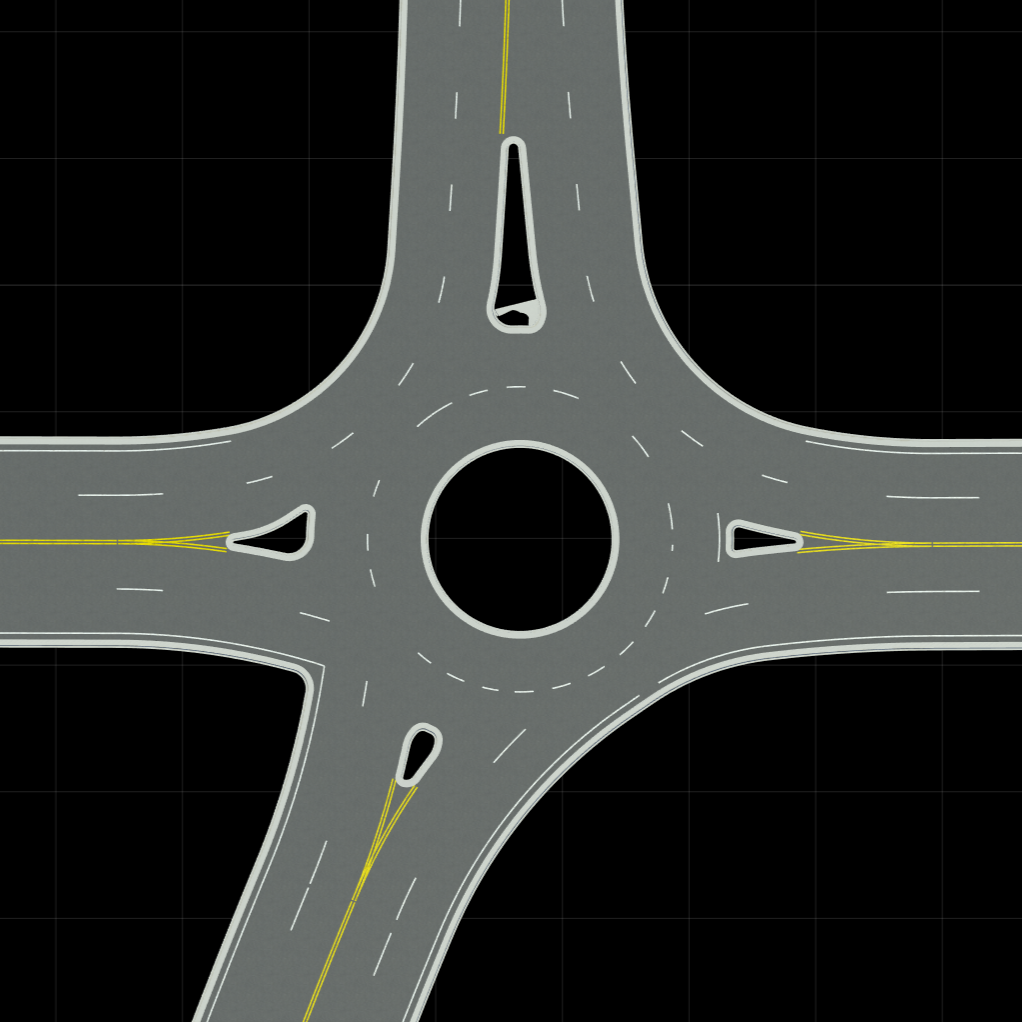}
         \caption{16m diameter}
         \label{fig:16mRIMG}
     \end{subfigure}
     \hfill
     \begin{subfigure}[b]{0.18\textwidth}
         \centering
         \includegraphics[width=\textwidth]{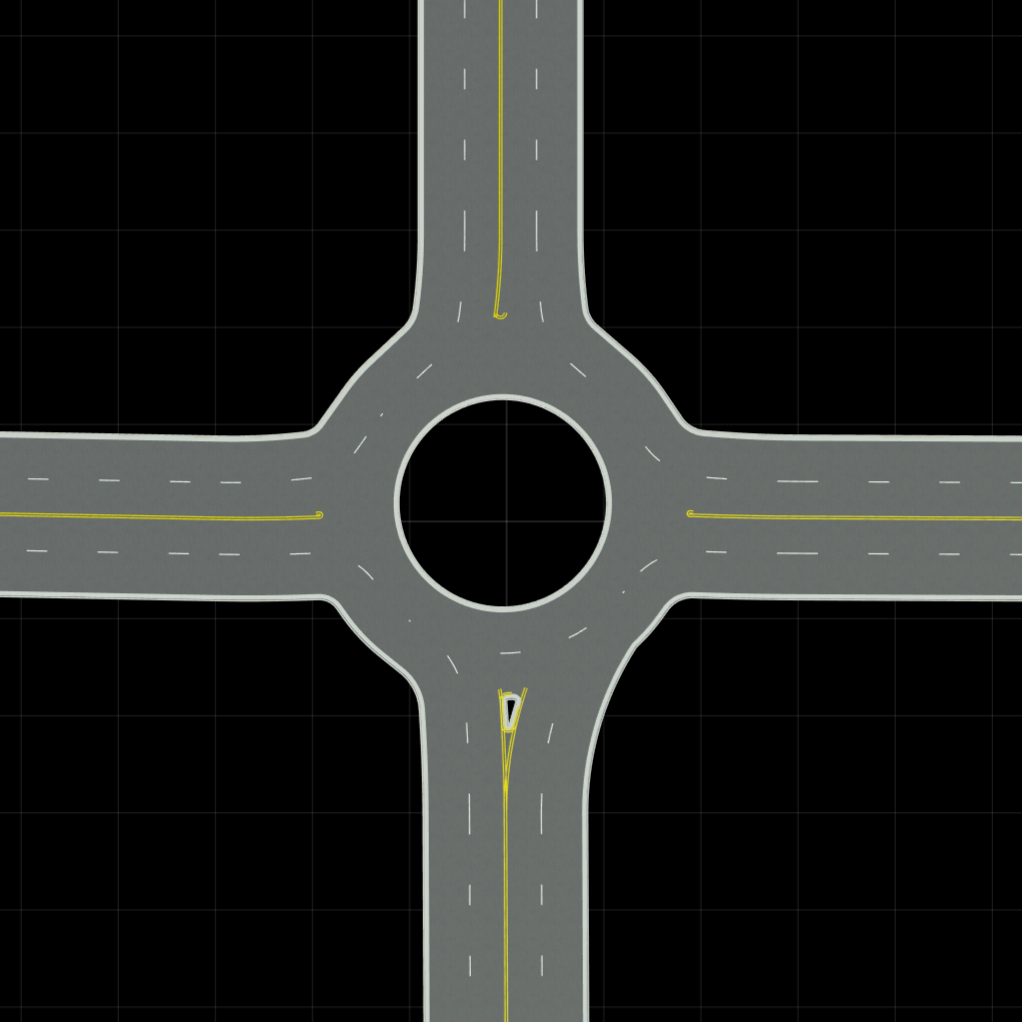}
         \caption{20m diameter}
         \label{fig:20mRIMG}
     \end{subfigure}
     \hfill
     \begin{subfigure}[b]{0.18\textwidth}
         \centering
         \includegraphics[width=\textwidth]{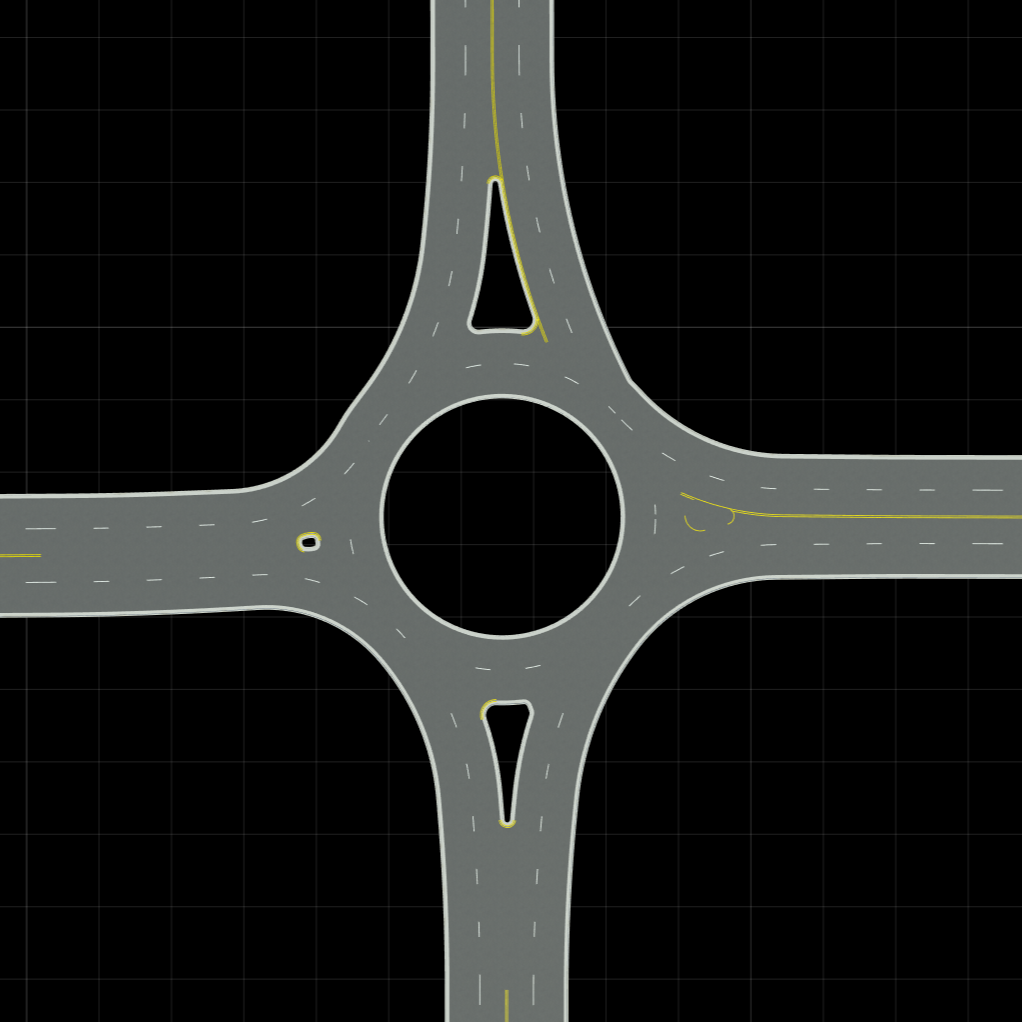}
         \caption{32m diameter}
         \label{fig:32mRIMG}
     \end{subfigure}
     \hfill
     \begin{subfigure}[b]{0.18\textwidth}
         \centering
         \includegraphics[width=\textwidth]{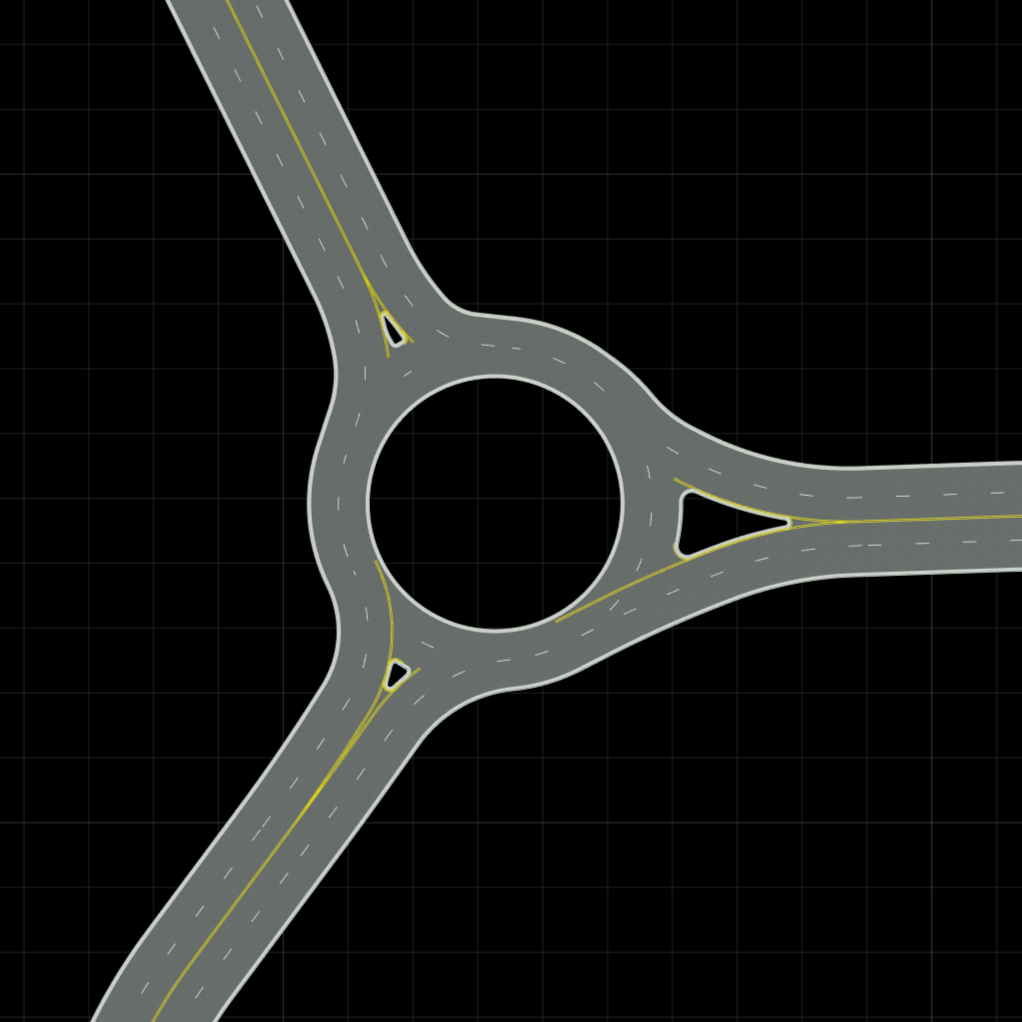}
         \caption{40m diameter}
         \label{fig:40mRIMG}
     \end{subfigure}
     \hfill
     \begin{subfigure}[b]{0.18\textwidth}
         \centering
         \includegraphics[width=\textwidth]{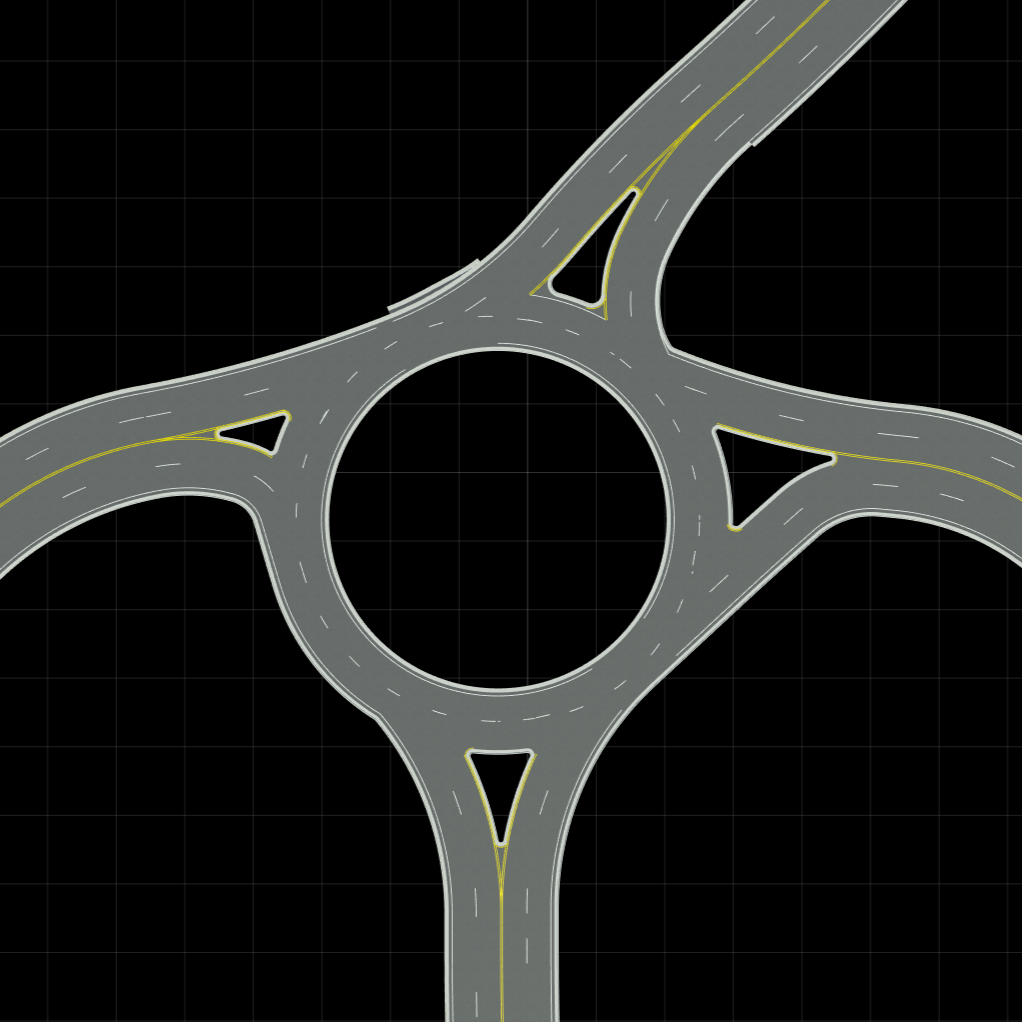}
         \caption{50m diameter}
         \label{fig:50mRIMG}
     \end{subfigure}
        \caption{Developed roundabout scenarios having different diameters}
        \label{fig:roundabouts}
\end{figure*} 

\subsection{Simulation environment}
Several open-source simulation environments have been developed for testing RL algorithms. Simulators used in autonomous driving research include, SVL \cite{rong_lgsvl_2020}, SUMO \cite{lopez_microscopic_2018}, and Carla \cite{dosovitskiy_carla_2017}. SVL lacks an active community which hinders its use. SUMO has been used for training the control function of an autonomous vehicle, but it is designed as a traffic simulator and therefore lacks realistic vehicle control, physics simulation for vehicle movement, and more complex perception inputs, such as Lidar.

Carla is an open-source autonomous vehicle driving simulator, supporting a large number of sensors, including RGB camera, Lidar, Radar, and inertial measurement unit (IMU). Traffic with different levels of aggressiveness can also be generated. Various vehicle properties, such as velocity, and sensor outputs, can be obtained through the Python API library, which also enables the control of such vehicle. Such API will be used along with the RLlib \cite{liang_rllib_2018} framework to train, test, and evaluate different RL algorithms in the Carla environment.

Furthermore, Carla computes and takes into consideration several physical properties. Due to the size and weight of tractor-trailer vehicles, several physical forces, not present in smaller vehicles, affect their movement. Firstly, the weight of the trailer can vary significantly, which in turn affects the acceleration and braking performance of the truck. Secondly, the high center of mass of the trailer can lead to rollover, where the trailer rolls on its side due to its speed and turning angle. Furthermore, Carla is also able to correctly simulate the articulated point joining the truck and trailer and how it affects both units. Such a high fidelity simulation software allows for the generation of quasi-real-life data, giving further confidence that the results obtained can be replicated in real-life scenarios.

Carla also allows the creation of custom vehicles and maps. Such a feature is essential to this research since, at the time of writing, no autonomous driving simulator offers a ready-made tractor-trailer model and suitable roundabout scenarios.

\subsection{Development of custom tractor-trailer model}
There are several steps in building a custom vehicle in Carla. The developed truck and trailer model used in this research is based on the work of Engles \footnote{https://github.com/frankeng/CarlaSemiTruckTrailer}. Despite having both the truck and trailer visual and physics models, improvements to the physics models were necessary to mimic real-life behaviour more accurately. 
Using Unreal Engine  \footnote{https://www.unrealengine.com/en-US}, the simulation development software used by Carla, vehicle meshes generated using Blender \footnote{https://www.blender.org/}, were integrated into the models to allow for correct detection by the collision sensors.

Furthermore, the correct engine capacity, which were implemented while also refining the articulated joint control between the truck and trailer.

\subsection{Development of custom scenarios}
\label{section:roundabout_making}
At the time of writing, Carla only offers two roundabouts scenarios. One roundabout is a single-lane roundabout with one exit, at the same point of entry, while the other roundabout is a two-lane roundabout with four exits. In order to correctly evaluate the developed model, roundabouts with different radii, number and position of exits, and number of lanes, need to be utilised. RoadRunner \cite{matlab_version_2022} is a 3D design tool used to create scenes for driving simulation. Such scenes can then be imported and used in Carla.

As illustrated in Figure \ref{fig:roundabouts}, five different roundabouts have been developed, having a diameter of 16m, 20m, 32m, 40m, and 50m. The 40m roundabout has 3 entry-exit points while the rest have 4 entry-exit points with each roundabout having different placement and angle of exits. Varying such properties presents different challenges when navigating truck and trailer vehicles, forcing a wider path when turning around small radii to avoid collisions. This wider path forces the truck and trailer to drive on other lanes, deviating from the original path. In the developed roundabouts, this may be necessary for both manoeuvring around the roundabout and for entering or exiting the roundabout. Each roundabout is approached by two, 3.7 meter wide lanes. These connect onto the two-lane roundabout. This allows for a total of 5 possible routes per entry. Each route is defined as a list of waypoints at the centre of the appropriate lane. No modifications were made to the waypoints where tighter radii occur.

As mentioned above, larger radii roundabouts are also developed. In such roundabouts, the truck and trailer vehicle can follow the conventional path and stay within its lane due to the low turning angle. Having such roundabouts improves the generality of the model by providing scenarios where no deviation from the path is required.

\subsection{Reinforcement Learning}
\subsubsection{Markov decision Process}


We define our quasi-end-to-end autonomous driving problem as a Markov decision process in an agent-environment interaction, formally described as a 5-tuple \(<S, A, R, f, \gamma>\), where the agent's goal is to find an optimal policy \(\pi: S \times A \rightarrow [0,1]\) which maximises the reward. The state space, \(S\), represents information obtained from the environment. The possible actions are defined in action space, \(A\). Applying the selected action to the environment results in a numerical reward, \(R\). The state-transition model, \(f\), defines the effects of actions on state changes while the discount factor, \(\gamma\), determines the degree of future rewards to take into consideration when evaluating actions \cite{Sutton2018}.

Our environment also possesses the Markov property, which is required for a Markov decision process. This is where the next state is only dependent on the current state and action
\cite{bremaud_markov_2020}. 


\subsubsection{Soft Actor-Critic}




Model-free deep RL methods suffer from two main problems. They require a large number of samples and are highly sensitive to the chosen hyperparameters. SAC aims to solve these issues by maximising the expected reward and policy entropy, ``succeeding at the task while acting as randomly as possible'' \cite{haarnoja_soft_2018}. This produces improved stability while achieving state-of-the-art performance.

By introducing policy entropy, the RL optimal policy can be defined as Equation \ref{eq:sac_optimal_policy}.

\begin{equation}
\label{eq:sac_optimal_policy}
\pi^{*} = \arg \max_{\pi} \mathop{\mathbb{E}}_{\tau \sim \pi} \biggr[\sum_{t=0}^{\infty}\gamma^{t}\Bigl(R(s_t,a_t,s_{t+1}) + \alpha H(\pi(.|s_t))\Bigl)\biggr]
\end{equation}

Where $H(\pi(.|s_t))$ is the entropy for a state $s_t$. The expected return is now made up of 2 components, the current reward and entropy, which is proportional to the non-negative parameter $\alpha$.

In twin-q SAC, the algorithm uses three networks, one policy network and two Q-function networks. The policy network is the probability that action \(a\) is chosen at time \(t\) given state \(s\) while the Q-function network is the expected return when choosing action \(a\) while in state \(s\). Having two independent Q-functions reduces the probability of overestimating \cite{fujimoto_addressing_2018}. Being an off-policy algorithm, SAC also makes use of a prioritised experience replay buffer filled with previous experiences. These experiences are used to train the current policy, selected based on their priority, leading to faster convergence \cite{wang_boosting_2019}.

\subsubsection{State space}
The features used to describe the state of the environment to the agent can be split up into 4 categories, having a total of continuous 69 elements. 
Firstly, in order for the agent to understand its distance away from the road boundary, the truck is equipped with 13 distance sensors at $15\degree$ intervals, originating from its centre, while the trailer has 8 equally spaced sensors on each side. This is shown in Figure \ref{fig:overview}.

Secondly, for the agent to understand its expected future position and heading, the angle between the vehicle's forward vector and the forward vector of future waypoints is obtained. This angle is calculated for 5 different waypoints, 1,2,5,7, and 10 waypoints ahead. Due to the articulated joint, the trailer's angle can widely vary from that of the truck, and therefore, these angles are calculated for both units. The angle between the truck and trailer is also provided for the agent to understand the relationship between them. Furthermore, the agent is also given the truck's angle to the centre of the lane computed by calculating the angle between the vectors produced from the previous waypoint to the truck, and from the previous waypoint to a future waypoint. 5 different waypoints are used to obtain a better picture of the expected future position. 

To represent the curvature of the route, not the path the tractor-trailer needs to take, the angle subtended by two equally distanced waypoints from a central waypoint is calculated. This is performed at 4 different distances, 5,7,10, and 12 waypoints, between each central and end waypoints, leading to a better understanding of the current route.
Taking the current waypoint as the central waypoint generates a value which describes the current curvature of the route. On the other hand, using the current waypoint as one of the end waypoints, with the two other waypoints being further along in the route, obtains a value which represents the future curvature of the route. Both of these values are essential for the agent to understand the current route and to be able to choose the appropriate action. Further to this, the radius of the route is approximated at 10 different intervals by generating a chord between every 5 waypoints and calculating their bisecting line. The intersection between two bisecting lines of adjacent chords, can be taken as the centre of a circle whose radius can be calculated.

Other states include the truck's forward velocity, and the hypotenuse distance to the next two waypoints. When taking tighter radii turns, the truck has to move away from the route and therefore a line at a right angle to the forward vector of a waypoint, can be used to inform the agent that despite not moving towards the waypoint, the general direction is correct, moving closer to the end of the route. The closest distance to this line, 1 and 2 waypoints ahead, is also added to the state space.
    
For easier interpretation by the RL algorithm, values are normalised between 0 and 1 while all angles are in radians.

\subsubsection{Action space}
A discrete action space is used to control the tractor-trailer vehicle and is composed of 9 actions. These actions differ in their steering angle values, starting from 0 and increasing at intervals of 0.2, both positively and negatively. The acceleration value is controlled by a proportional–integral–derivative (PID) controller, which maintains a forward velocity of $30km/h$. This velocity was observed to be sufficient in completing the episode in a suitable timeframe while also being slow enough to complete the task correctly. The PID was tuned using the Ziegler-Nichols tuning method \cite{ziegler_optimum_1993}.

\subsubsection{Reward function}
Our reward function is composed of a utility function and a reward shaping function. As mentioned by Knox et al. \cite{knox_reward_2023}, reward shaping should be kept to a minimum and be justified since it can lead to undesirable behaviour, decreasing overall performance, despite its intention being the opposite. For the utility function, a +100 reward is given for each waypoint passed. A -500 reward is given if the tractor-trailer vehicle collides with the kerb, diverges too far from the route, or the number of time steps in that particular episode is too large.

The distance between the truck and the centre of the lane is calculated and a negative reward, calculated by Equations \ref{distanceEQ1} and \ref{distanceEQ2}, is given.

\begin{eqnarray}
\label{distanceEQ1}
r_d & = & clip(distance\_to\_center\_of\_lane, 0,4)/4\\
\label{distanceEQ2}
r_d & = & r_d \times -1.5
\end{eqnarray}

Such a reward is observed in several previous research and is of high importance in our work since the tractor-trailer vehicle may opt to move in between lanes to avoid collision in all occasions. This reward pushes the agent to drive closer to the centre of the lane when possible. The 1.5 times increase was obtained through trial and error, which outputs the appropriate weighted reward value when compared to the other rewards. This reward is considered as part of the reward shaping function.

\subsubsection{Training Setup}

Training was performed using the RLlib framework and the twin-q SAC algorithm. A total of 52 routes were used during training, which were randomly chosen from the 16m, 32m and 50m roundabouts defined in Section \ref{section:roundabout_making}. The model was trained for 1.5 million steps equating to 5700 episodes, using 8 simulators in parallel, where it converged to an optimal mean reward. Table \ref{tab:hyperparameters} illustrates the hyperparameters used.

\begin{table}[]
\begin{center}
\caption{Hyperparameters for SAC algorithm}
\label{tab:hyperparameters}
\begin{tabular}{ll}
\hline

\textbf{Hyperparameters}          & \textbf{Values}    \\ \hline
gamma                             & 1                  \\
n\_step                           & 1                  \\
batch size                        & 256                \\
reply buffer capacity             & 600000             \\
twin q                            & True               \\
exploration algorithm             & Epsilon Greedy     \\
initial epsilon                   & 1.0                \\
final epsilon                     & 0.01               \\
exploration epsilon timesteps     & 1000000            \\
q model hidden layers      & {[}512,512,1024{]} \\
policy model hidden layers & {[}512,512,1024{]} \\
\hline
\end{tabular}
\end{center}
\end{table}

\section{Experimental Results and Evaluation}
\label{chp:experimental_results}
In this section, the experimental results will be discussed and evaluated. Testing was performed on the 20m and 40m roundabouts. These were not used during training and have different physical properties. This allows for better evaluation and can more accurately indicate the generality of the model. The testing routes include both larger and smaller radii turns. A total of 23 different routes were tested each ran 30 times, resulting in 690 episodes. The routes were analysed and marked whether or not the tractor-trailer vehicle had to move away from the centre of the lane in order to successfully complete the route.

\subsection{Quantitative analysis}

\begin{figure}
\centerline{\includegraphics[width=0.45\textwidth]{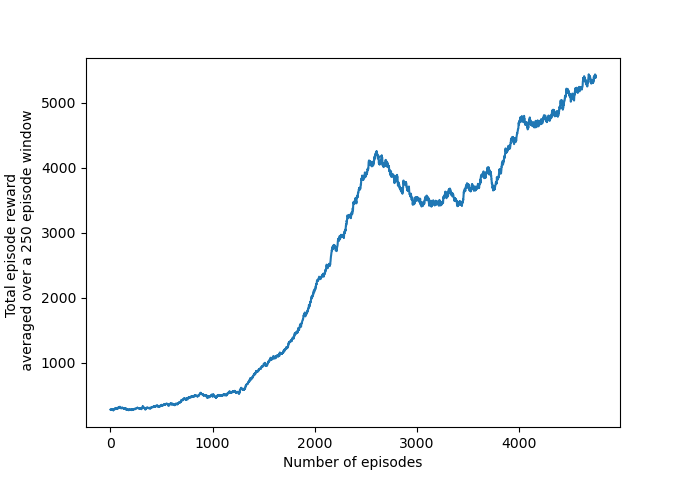}}
\caption{Episode reward mean averaged over a 250 sliding window} \label{img:reward_mean}
\end{figure}

\begin{figure}
\centerline{\includegraphics[width=0.45\textwidth]{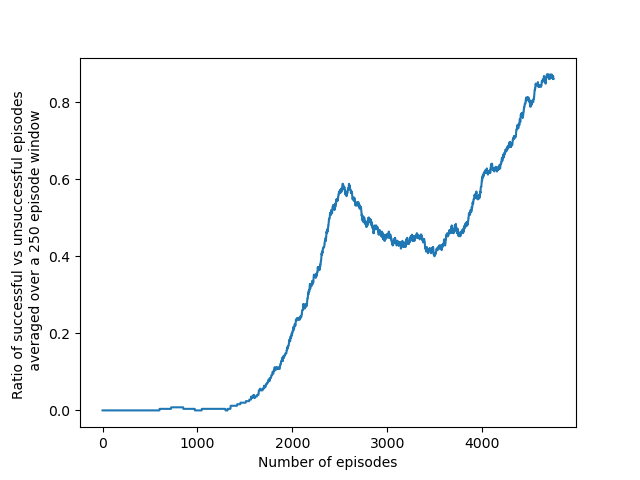}}
\caption{Ratio of successful to unsuccessful episodes completion averaged over a 250 sliding window} \label{img:succesrate}
\end{figure}

Figure \ref{img:reward_mean} illustrates the episode reward obtained during training, while Figure \ref{img:succesrate} shows the ratio of successful to unsuccessful episodes during training. Initially, both the reward and success rates gradually increase. After around 2500 episodes, the episode reward is observed to decrease alongside a decrease in success rate. Evaluating the model at this stage illustrates that the policy started to explore different actions for tight radius paths where, up to now, such paths were always unsuccessful. After 3700 episodes, the episode reward and success rates significantly increase indicating that the policy has started to successfully navigate through both small and large radius paths. At the end of training, the episode reward is observed to plateau at around 5100, while the success rate rises to just above 0.8. 


When using the developed policy on the training routes, a 79\% success rate was achieved on the training data. 20\% were unsuccessful due to the trailer colliding with the kerb while turning around the roundabout. The remaining 1\%, 1 route, failed due to the truck colliding with the kerb. This was observed to occur when the agent failed to move away from the kerb after being spawned close to it.

Using roundabout scenarios with different physical properties than those used during training, a testing success rate of 73\% is obtained, with the remaining 17\% of episodes all failing due to a collision with the trailer. It should be noted that no episode timed out due to taking too long to complete. This is mainly because the acceleration is a constant non-zero value and therefore the vehicle always moves forward. In our case, an episode could have timed out when the agent chose to take a different, longer path, but this never occurred.

\begin{figure}
     \centering
     \begin{subfigure}[b]{0.22\textwidth}
         \centering
         \includegraphics[width=\textwidth]{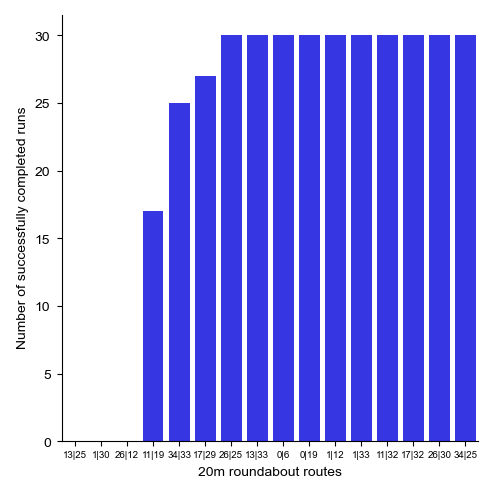}
         \caption{20m roundabout}
         \label{fig:20mcompelted}
     \end{subfigure}
     \hfill
     \begin{subfigure}[b]{0.22\textwidth}
         \centering
         \includegraphics[width=\textwidth]{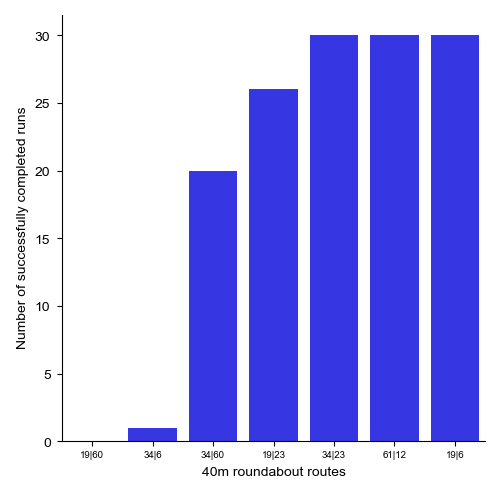}
         \caption{40m roundabout}
         \label{fig:40mcompleted}
     \end{subfigure}
     \hfill
        \caption{Success rate of each testing route}
        \label{fig:completed}
\end{figure}


Analysing the data shows that some routes were almost always completed successfully, while others were never completed successfully. Figure \ref{fig:20mcompelted} illustrates this for the 20m roundabout, where all the unsuccessful routes were ones which needed to exit at the second exit. The trailer collided either with the central kerb or the outer kerb of the roundabout. Since these routes, which exit 2 entry-exit points away from the start, are quite straight, the agent would have incorrectly determined that less deviation from the centre of the lane is required for such straight routes.

Figure \ref{fig:40mcompleted} visualises the success rate per route for the 40m diameter roundabout. The failing cases are observed to be the trailer colliding with the central kerb of the roundabout, where despite the truck starting the deviate away from the centre of the lane, the deviation is not large enough, failing to allow enough space for the trailer to follow a tighter radius.

Being trained on the 16m, 32m, and 50m diameter roundabouts, the generated model may have not been exposed to enough variance in path curvature.

The distance to the centre of the lane is also be used as a metric to determine how well the tractor-trailer vehicle performs. Lower values indicate more human-like behaviour, but this can only be expected when the route does not contain low radius turns. The mean distance to the centre of the lane for the routes which did not require the truck to move out of the lane, was found to be 0.73m. Considering the 3.7m wide lanes used and the truck being 2.5m wide, such a value is deemed acceptable, since on average, the truck stays within its lane when possible. 

\begin{figure}
     \centering
     \begin{subfigure}[b]{0.2\textwidth}
         \centering
         \includegraphics[width=\textwidth]{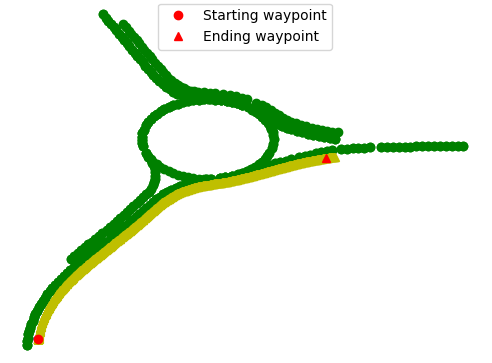}
         \caption{Route in context of the whole roundabout}
         \label{fig:2766-1}
     \end{subfigure}
     \hfill
     \begin{subfigure}[b]{0.2\textwidth}
         \centering
         \includegraphics[width=\textwidth]{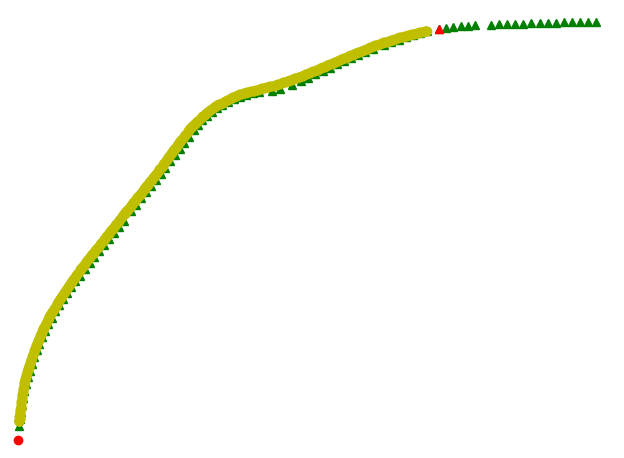}
         \caption{Close up of the route}
         \label{fig:2766-2}
     \end{subfigure}
     \hfill
        \caption{Simple route completed successfully with the tractor-trailer vehicle close to the centre of the lane}
        \label{fig:2766}
\end{figure}

\begin{figure}
     \centering
     \begin{subfigure}[b]{0.2\textwidth}
         \centering
         \includegraphics[width=\textwidth]{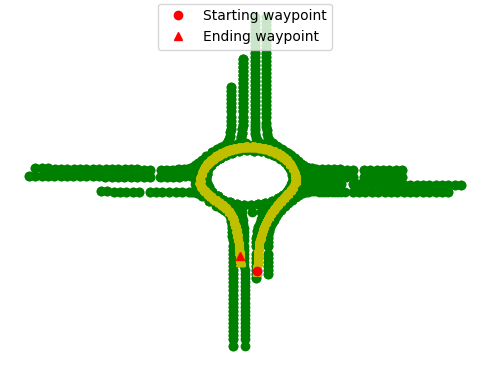}
         \caption{Route in context of the whole roundabout}
         \label{fig:344-1}
     \end{subfigure}
     \hfill
     \begin{subfigure}[b]{0.2\textwidth}
         \centering
         \includegraphics[width=\textwidth]{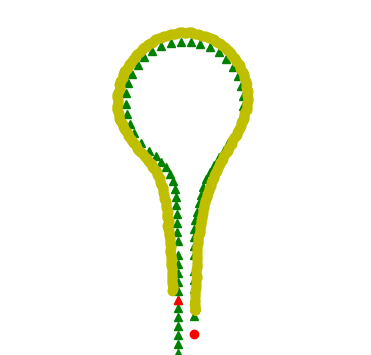}
         \caption{Close up of the route}
         \label{fig:344-2}
     \end{subfigure}
     \hfill
        \caption{Difficult route completed successfully with the tractor-trailer vehicle as close to the centre of the lane as possible}
        \label{fig:344}
\end{figure}

In Figures \ref{fig:2766} and \ref{fig:344}, the green points represent the route provided to the agent, while the yellow points represent the path taken by the truck. The red circle represents the starting point, while the red triangle represents the end point.

Figure \ref{fig:2766} illustrates the path taken by the tractor-trailer vehicle when navigating a route which does not require driving away from the normal route due to its large curvature. As can be seen in Figure \ref{fig:2766-2}, the vehicle very closely follows the route even when a slight curvature is encountered, indicating human-like behaviour. Noticing the green points to the left of the yellow points in Figure \ref{fig:2766-1}, one can conclude that despite the vehicle having plenty of space to drive further away from the right wall, the agent correctly follows the given path. 

Figure \ref{fig:344} shows the path taken by the agent to manoeuvre through a route with a low radius. Following the given route, the centre of the lane, would result in the trailer colliding with the side of the road when turning since the trailer's rear wheels follow a tighter radius path than the front wheels of the truck. In order to avoid this, the agent must anticipate that a low radius section is expected and move away from the centre of the lane before starting to turn. This is shown to happen in Figure \ref{fig:344-2} after passing the first quarter of the roundabout, where the yellow path of the truck starts to deviate from the given green route. Following this, to avoid a collision between the trailer and the kerb, the agent must continue to drive off-centre, allowing the trailer enough space to turn at a tighter radius. Figure \ref{fig:344-2} illustrates this behaviour from the first quarter of the roundabout to the final quarter of the roundabout. 

\begin{figure}
\centerline{\includegraphics[width=0.48\textwidth]{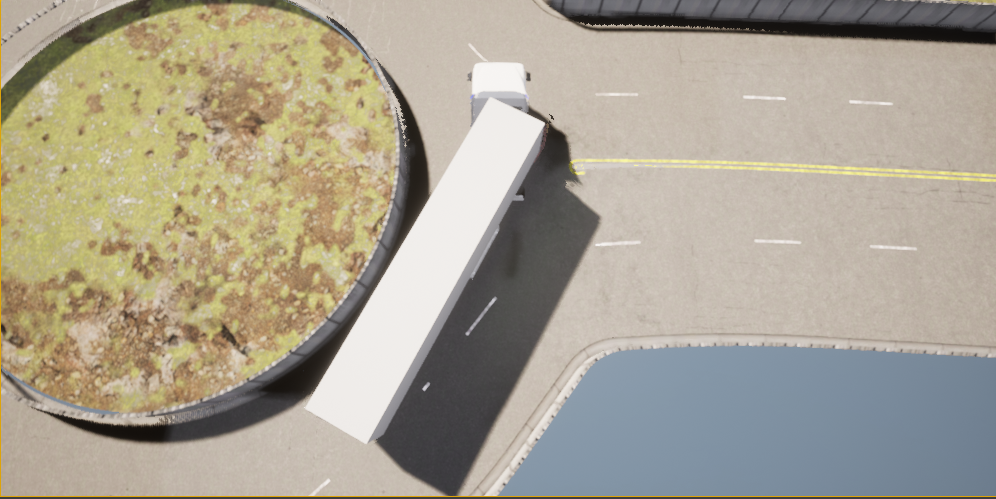}}
\caption{Tractor-trailer vehicle manoeuvring through a low radius curve} \label{img:truck_diff_route}
\end{figure}

Figure \ref{img:truck_diff_route} is a visual from the Carla simulation showing the tractor-trailer vehicle halfway through this route. As can be seen, the truck is driving in the centre of the two lanes in order to allow the trailer enough space not to collide with the kerb. It should also be noted that the trailer is at the centre of the original lane, showing that the truck deviates as little as possible in order to complete the episode, resembling human-like behaviour with the least intrusion possible into other lanes. This is further shown in Figure \ref{fig:344-1} where the green points to the right of the path confirm that the agent had plenty of space to drive further away from the roundabout, leading to an easier path but opted to stay as close to the original lane as possible. Similarly, in Figure \ref{fig:344-2} one can also notice that, when feasible, the truck is driven as close to the centre of the lane as possible. A slight deviation from the centre of the lane is present at the end of the route.

\section{Conclusion and Future Work}
\label{chp:conc}

In this work, we developed a quasi-end-to-end system for the autonomous navigation of a tractor-trailer vehicle through a roundabout intersection. Being an articulated vehicle, the rear trailer wheels follow a tighter radius path than the front wheels of the truck and therefore the truck has to deviate from the centre of the lane to avoid the trailer colliding with the roundabout kerb. Using steering actions, a PID controller to maintain a constant velocity, and a twin-q SAC algorithm, a 73\% success rate was achieved. Testing on roundabouts with different physical properties than those used for training, the developed model achieved human-like behaviour, with the truck being as close to the centre of the lane as possible. Furthermore, when tasked to solve tighter radius routes, the agent correctly perceives this and sufficiently deviates from the centre of the lane.

At the time of writing, such work is the first-of-its-kind, limiting our ability to compare to state-of-the-art solutions. Future research can therefore take several different paths. Sharper turns, such as $90\degree$ and U-turns, can be tested to further develop this field. Introducing traffic would test the correct cooperation with other vehicles which may have different physical properties, therefore behaving differently. Furthermore, since such articulated vehicles may occupy more than one lane, additional awareness of surrounding vehicles is required.


\bibliography{ecai-sample-and-instructions}
\end{document}